\documentclass[conference]{IEEEtran}

\usepackage[cmex10]{amsmath}
\usepackage{amsthm}
\usepackage{amssymb}
\usepackage{mathrsfs}
\usepackage{graphicx}
\usepackage{float}
\usepackage{array}
\usepackage{epstopdf}
\usepackage{multirow}

\begin{document}

\title{DeepIris: Iris Recognition Using A Deep Learning Approach}

\author{Shervin Minaee$^*$, Amirali Abdolrashidi$^{\dagger}$  \\
$^*$New York University
\\ $^{\dagger}$University of California, Riverside\\ \\
}

\maketitle

\begin{abstract}
Iris recognition has been an active research area during last few decades, because of its wide applications in security, from airports to homeland security border control. 
Different features and algorithms have been proposed for iris recognition in the past.
In this paper, we propose an end-to-end deep learning framework for iris recognition based on residual convolutional neural network (CNN), which can jointly learn the feature representation and perform recognition.
We train our model on a well-known iris  recognition dataset using only a few training images from each class, and show promising results and improvements over previous approaches. 
We also present a visualization technique which is able to detect the important areas in iris images which can mostly impact the recognition results. 
We believe this framework can be widely used for other biometrics recognition tasks, helping to have a more scalable and accurate systems.
\end{abstract}

\IEEEpeerreviewmaketitle

\section{Introduction}
To personalize an experience or make an application more secure and less accessible to undesired people, we need to be
able to distinguish a person from everyone else. 
There are various ways to identify a person, and biometrics are one of the most secure options so far. 
They can be divided into two categories: behavioral and physiological features. Behavioral features are those actions that a per-son can uniquely create or express, such as signatures, walking rhythm, and the physiological features are those characteristics that a person possesses, such as fingerprints and iris pattern. Many works revolved around recognition and categorization of such data including, but not limited to, fingerprints,faces, palmprints and iris patterns \cite{biomet1}-\cite{biomet5}.

Iris recognition systems are widely used for security applications, since they contain a rich set of features and do not
change significantly over time. They are also virtually impossible to fake. One of the first modern algorithms for iris
recognition was developed by John Daugman and used 2D Gabor wavelet transform [6].
Since then, there have been various works proposing different approaches for iris recognition. 
Many of the traditional approaches follow the two-step machine learning approach, where in the first step a set of hand-crafted features are derived from iris images, and in the second step a classifier is used recognize the iris images.
Here we will discuss about some of the previous works proposed for iris recognition.

In a more recent work, Kumar \cite{prev1} proposed an algorithm based on a combination of Log-Gabor, Haar wavelet, DCT and FFT features, and achieved  high accuracy. 
In \cite{prev2}, Farouk proposed an scheme which uses elastic graph matching and Gabor wavelet for iris recognition. Each iris is represented as a labeled graph and a similarity function is defined to compare the two graphs. 
In \cite{prev3}, Belcher used region-based SIFT
descriptor for iris recognition and achieved a relatively good performance. 
In \cite{prev4}, Umer proposed an algorithm for iris recognition using multiscale morphologic features.
More recently, Minaee et al \cite{prev5} proposed an iris recognition using multi-layer scattering convolutional networks, which decomposes iris images using Wavelets of different scales and orientations, and used those features for iris recognition. An illustration of the decomposed images in the first two layers of scattering network is shown in Fig 1.
\begin{figure}[h]
\begin{center}
    \includegraphics [scale=0.4] {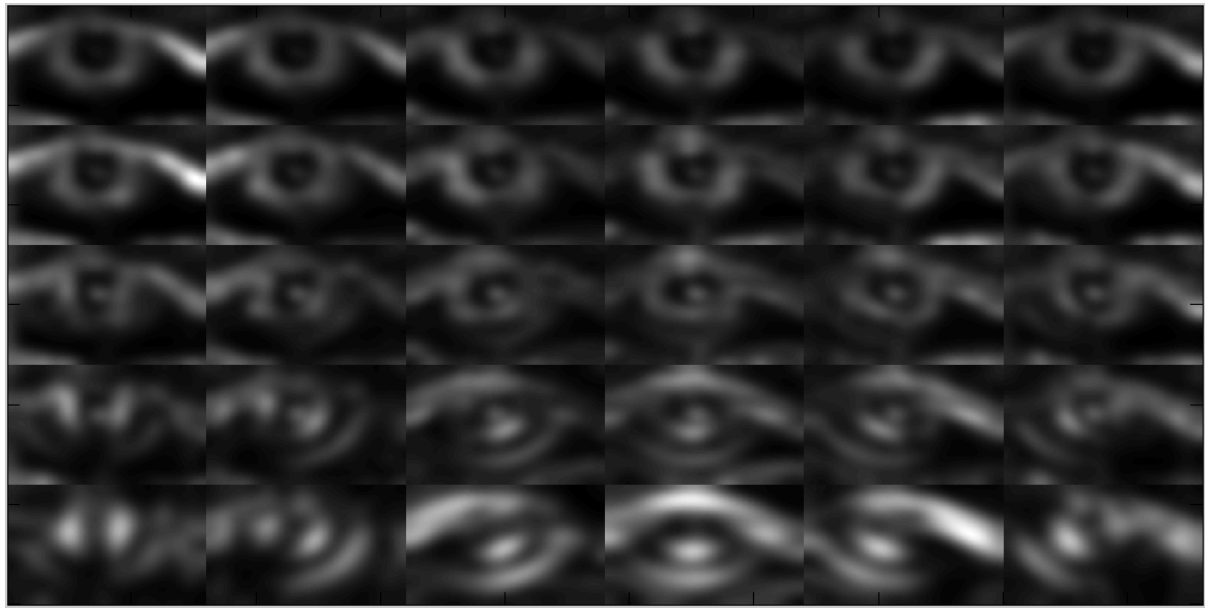}
    \includegraphics [scale=0.4]  {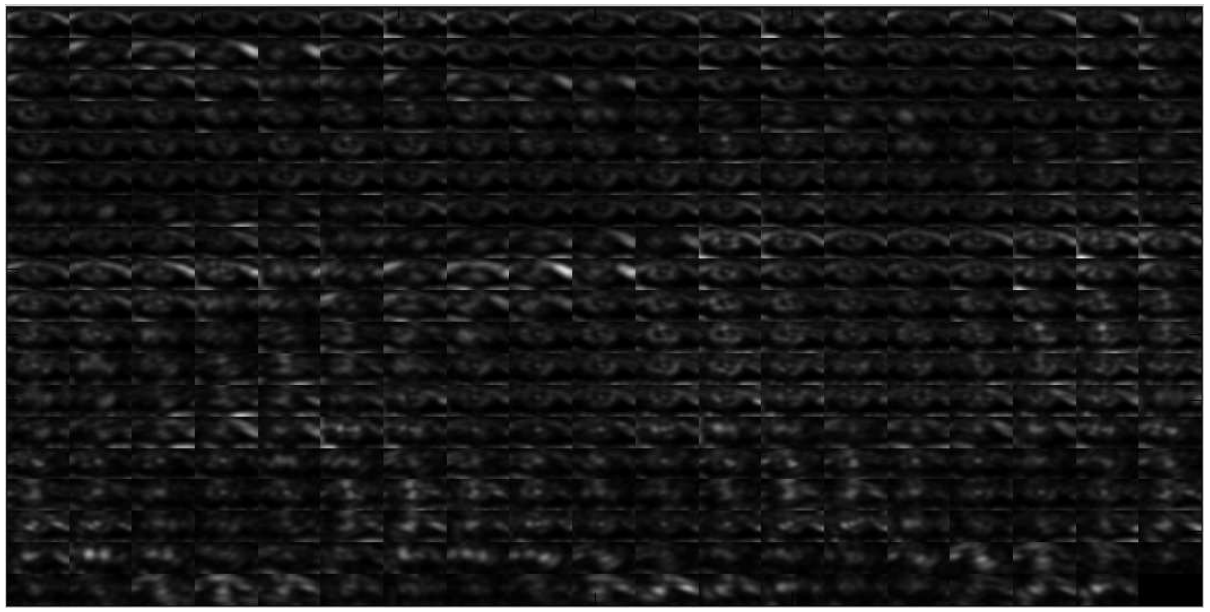}
\end{center}
  \caption{The images from the first (on top) and second layers of scattering transform  \cite{prev5} for a sample iris image. Each image is capturing the wavelet energies along specific orientation and scale.}
\end{figure}

Although many of the previous works for iris recognition achieve high accuracy rates, they involve a lot of pre-processing (including iris segmentation, and unwraping the original iris into a rectangular area) and using some hand-crafted features, which may not be optimum for different iris datasets (collected under different lightning and environmental conditions).
In recent years, there have been a lot of focus on developing models for jointly learning the features, while doing prediction. 
Along this direction, convolutional neural networks \cite{cnn} have been very successful in various computer vision and natural language processing (NLP) tasks \cite{alexnet}.
Their success is mainly due to three key factors: the availability of large-scale manually labeled datasets, powerful processing tools (such Nvidia's GPUs), and good regularization techniques (such as dropout, etc) that can prevent overfitting problem.

Deep learning have been used for various problems in computer vision, such as image classification, image segmentation, super-resolution, image captioning, emotion analysis, face recognition, and object detection,  and significantly improved the performance over traditional approaches \cite{cnn1}-\cite{cnn8}. 
It has also been used heavily for various NLP tasks, such as sentiment analysis, machine translation, name-entity-recognition, and question answering \cite{nlp1}-\cite{nlp5}.

More interestingly, it is shown that the features learned from some of these deep architectures can be transferred to other tasks very well. In other words, one can get the features from a trained model for a specific task and use it for a different task (by training a classifier/predictor on top of it) \cite{offshelf}. 
Inspired by \cite{offshelf}, Minaee et al. \cite{iriscnn} explored the application of learned convolutional features for iris recognition and showed that features learned by training a ConvNet on a general image classification task, can be directly used for iris recognition, beating all the previous approaches.

For iris recognition task, there are several public datasets with a reasonable number of samples, but for most of them the number of samples per class is  limited, which makes it difficult to train a convolutional neural network from scratch on these datasets. 
In this work we propose a deep learning framework for iris recognition for the case where only a few samples are available for each class (few shots learning).

The structure of the rest of this paper is as follows. Section II provides the description of the overall proposed framework. Section III  provides the experimental studies and comparison with previous works.
And finally the paper is concluded in Section IV.

\section{The Proposed Framework}
In this work we propose an iris recognition framework based on transfer learning approach. 
We fine-tune a pre-trained convolutional neural network (trained on ImageNet), on a popular iris recognition dataset. 
Before discussing about the model architecture, we will provide a quick introduction of transfer learning. 
%By initializing the model weights as the pre-trained one on ImageNet, we will start from a network state which has a good understanding of general image description.

\subsection{Transfer Learning}
Transfer learning is a machine learning technique in which a model trained on one task is modified and applied to another related task, usually by some adaptation toward the new task.
For example, one can imagine using an image classification model trained on ImageNet \cite{imagenet} to perform medical image classification. 
Given the fact that a model trained on general purpose object classification should learn an abstract representation for images, it makes sense to use the representation learned by that model for a different task. 
%There have been several promising works based on pre-trained deep learning models to perform a different task in the past few years. 
%Transfer learning is mainly useful for tasks where enough training samples are not available to train a model from scratch, such as medical image classification for rare diseases in which enough labeled samples are not available. This is especially the case for models based on deep neural networks, which have a large number of parameters to train. Figure \ref{fig:VGG16} shows the block-diagram of a sample transfer learning approach for histology images based on a pre-trained CNN model.

There are two main ways in which the pre-trained model is used for a different task. 
In one approach, the pre-trained model is treated as a feature extractor, and then a classifier/regressor model is trained on top of that to perform the second task. In this approach the internal weights of the pre-trained model are not adapted to the new task. 
One can think of using a pre-trained language model for deriving word representation used in another task (such as sentiment analysis, NER, etc.) as an example of the first approach.
In the second approach, the whole network (or a subset of layers/parameters of the model) is fine-tuned on the new task, therefore the pre-trained model weights are treated as the initial values for the new task, and are updated during the training procedure.

\subsection{Iris Image Classification Using ResNet}
In this work, we focused on iris recognition task, and chose a dataset with a large number of subjects, but limited number of images per subject, and proposed a transfer learning approach to perform identity recognition using a deep residual convolutional network.
We use a pre-trained ResNet50 \cite{cnn1} model trained on ImageNet dataset, and fine-tune it on our training images. 
ResNet is popular CNN architecture which was the winner of ImageNet 2015 visual recognition competition. It generates easier gradient flow for more efficient training. 
The core idea of ResNet is introducing a so-called \textit{identity shortcut connection} that skips one or more layers, as shown in Figure \ref{fig:Resnet}.
This would help the network to provide a direct path to the very early layers in the network, making the gradient updates for those layers much easier.
\begin{figure}[h]
\begin{center}
   \includegraphics[width=0.7\linewidth]{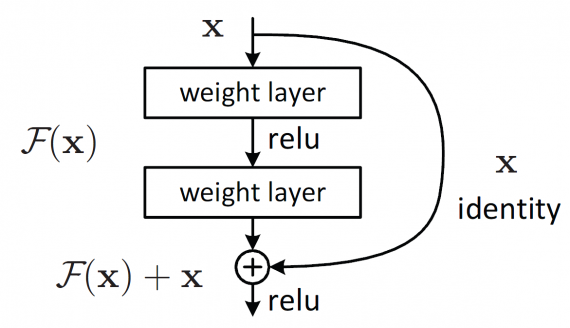}
\end{center}
   \caption{The residual block used in ResNet Model}
\label{fig:Resnet}
\end{figure}

To perform recognition on our iris dataset, we fine-tuned a ResNet model with 50 layers on the augmented training set.
The overall block-diagram of the ResNet50 model, and how it is used for iris recognition is illustrated in Figure 3.
\begin{figure*}[h]
\begin{center}
   \includegraphics[width=0.999\linewidth]{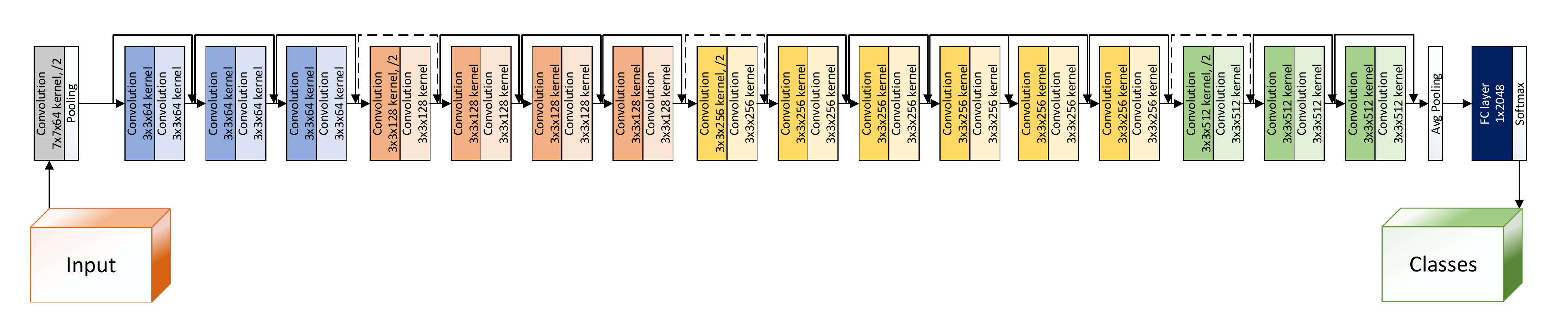}
\end{center}
   \caption{The architecture of ResNet50 neural network \cite{cnn1}, and how it is transferred for iris recognition. The last layer is changed to match the number of classes in our dataset.}
\label{fig:Resnet}
\end{figure*}

%We applied several data augmentation techniques to increase the number of training samples, including horizontal and vertical flip, random crops, and small distortions. By adding these augmentation, the number of training samples is increased by a factor of $\sim$3x.
We fine-tune this model for a fixed number of epochs, which is determined based on the performance on a validation set, and then evaluate it on the test set.
This model is then trained with a cross-entropy loss function. To reduce the chance of over-fitting the $\ell_2$ norm can be added to the loss function, resulting in an overall loss function as: 
\begin{equation}
\begin{aligned}
& \mathcal{L}_{final}=  \mathcal{L}_{class}+  \lambda_1 ||W_{fc}||_F^2
\end{aligned}
\end{equation}
where $\mathcal{L}_{class}= - \sum_{i} p_i \log(q_i)$ is the cross-entropy loss, and $||W_{fc}||_F^2$ denotes the Frobenius norm of the weight matrix in the last layer.
We can then minimize this loss function using stochastic gradient descent or Adam optimizer.

\section{Experimental Results}
In this section we provide the experimental results for the proposed algorithm, and the comparison with the previous works on this dataset.

Before presenting the result of the proposed model, let us first talk about the hyper-parameters used in our training procedure.
We train the proposed model for 100 epochs using an Nvidia Tesla GPU.
The batch size is set to 24, and Adam optimizer is used to optimize the loss function, with a learning rate of 0.0002.
All images are down-sampled to 224x224 before being fed to the neural network.
All our implementations are done in PyTorch \cite{pytorch}.
We present the details of the datasets used for our work in the next section, followed by  quantitative and visual  experimental results.

\subsection{Dataset}
We have tested our algorithm on the IIT Delhi iris database, which contains 2240
iris images captured from 224 different people. The resolution of these images is 320x240 pixels \cite{iit_delhi}.
Six sample images from this dataset are shown in Fig 4. 
As we can see the iris images in this dataset have slightly different color distribution, as well as different sizes.
\begin{figure}[h]
\begin{center}
    \includegraphics [scale=0.4] {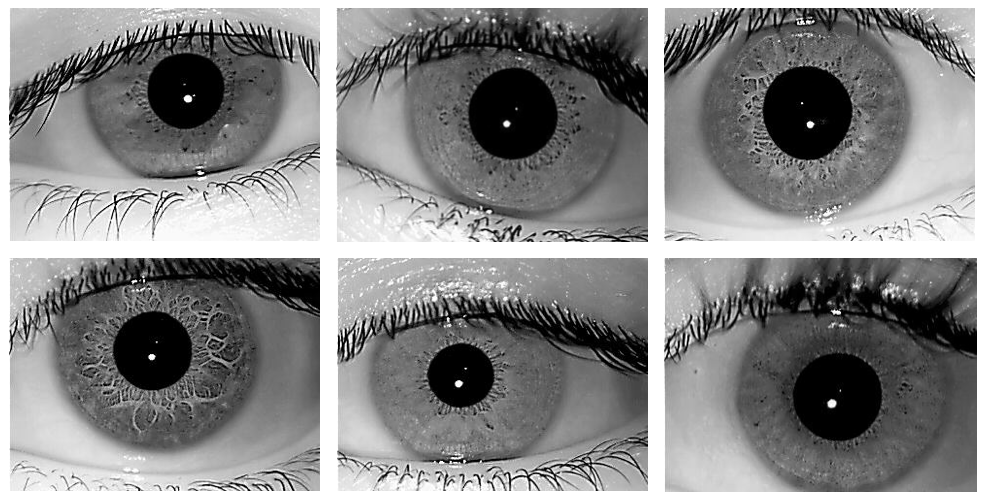}
\end{center}
  \caption{Six sample iris images from IIT Delhi dataset \cite{iit_delhi2}.}
\end{figure}

For each person, 4 images are used as test samples randomly, and the rest are using for training and validation.

\subsection{Recognition Accuracy}
Table I provides the recognition accuracy achieved by the proposed model and one of the previous works on this dataset, for iris identification task.
%It is worth to mention that the model in \cite{scat}, also uses features derived from convolutional neural network with pre-defined filters, which can essentially be though as a feature extractor by capturing higher order statistics.
\begin{table}[ht]
\centering
  \caption{Comparison of performance of different algorithms}
  \centering
\begin{tabular}{|m{4cm}|m{1.5cm}|}
\hline
Method  & Accuracy Rate\\
%\hline
%Direct Pore Matching \cite{pore} &   \ \ \ \ \ \ 79.51\% \\
%\hline
%Adaptive pore modeling \cite{adap_pore} &   \ \ \ \ \ \ 88.49\% \\
%\hline
%Scattering network \cite{scat} &   \ \ \ \ \ \ 92\% \\
\hline
Multiscale Morphologic Features \cite{prev4}  &   \ \ \ \ \ \ 87.94\% \\
\hline
 The proposed algorithm  &  \ \ \ \ \ \ 95.5\%\\
\hline
\end{tabular}
\label{TblComp}
\end{table}

\subsection{Important Regions Visualization} 
\begin{figure}[h]
\begin{center}
    \includegraphics [scale=0.5] {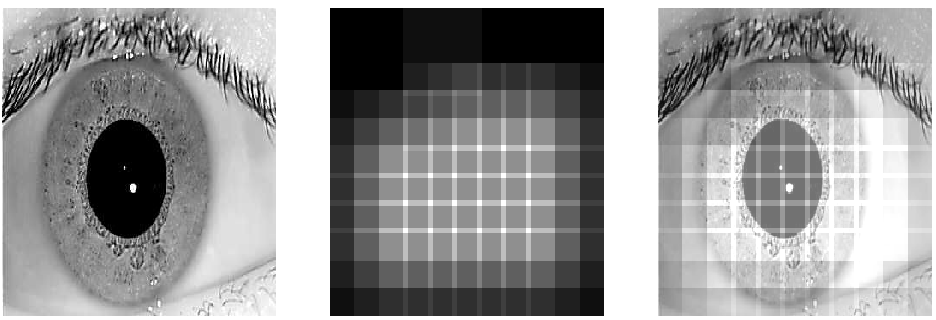}
    \includegraphics [scale=0.5] {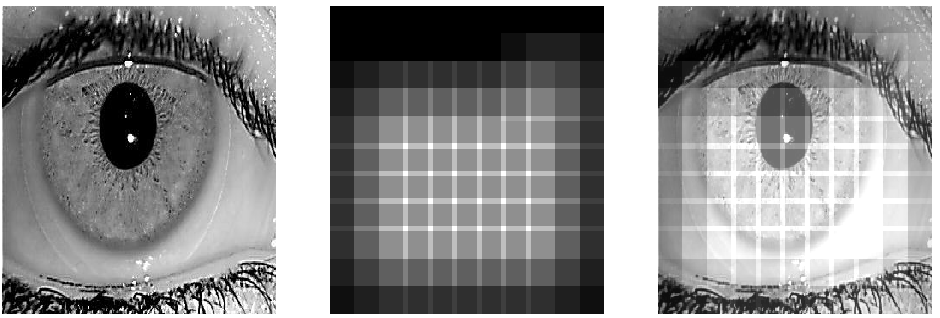}
    \includegraphics [scale=0.5] {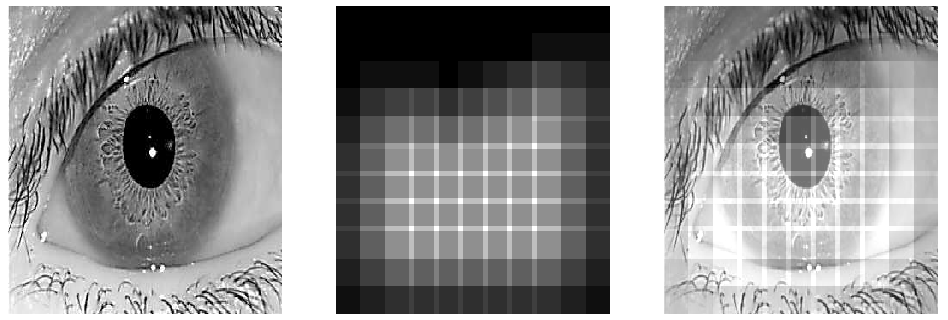}
    \includegraphics [scale=0.5] {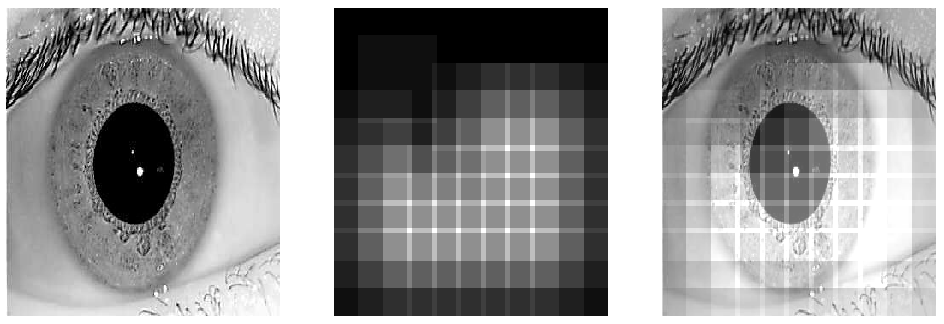}
\end{center}
  \caption{The saliency map of important regions for Iris recognition.}
\label{fig:finger_visual}
\end{figure}

Here we provide a simple approach to visualize the most important regions while performing iris recognition using convolutional network, inspired by the work in \cite{fergus}. 
We start from the top-left corner of an image, and each time zero out a square region of size $N$x$N$ inside the image, and make a prediction using the trained model on the occluded image.
If occluding that region makes the model to mis-label that iris image, that region would be considered as an important region, while doing iris recognition.
On the other hand, if removing that region does not impact the model's prediction, we infer that region is not as important.
Now if we repeat this procedure for different sliding windows of $N$x$N$, each time shifting them with a stride of $S$, we can get a saliency map for the most important regions in recognizing fingerprints.
The saliency maps for four example iris images are shown in Figure \ref{fig:finger_visual}.
As it can be seen, most regions inside the iris area seem to be important while doing iris recognition.

\section{Conclusion}
In this work we propose a deep learning framework for iris recognition, by fine-tuning a pre-trained convolutional model on ImageNet. 
This framework is applicable for other biometrics recognition problems, and is specially useful for the cases where there are only a few labeled images available for each class.
We apply the proposed framework on a well-known iris dataset, IIT-Delhi, and achieved promising results, which outperforms previous approaches on this datasets.
We train these models with very few original images per class.
We also present a visualization technique for detecting the most important regions while doing iris recognition.

\section*{Acknowledgment}
The authors would like to thank IIT Delhi for providing the iris dataset used in this work. We would also like to thank Facebook AI research for open sourcing the PyTorch package.

\end{document}